\documentclass[11pt, a4paper]{article}
\usepackage[utf8]{inputenc}

\usepackage{authblk}
\usepackage{lettrine}
\usepackage[margin=1in]{geometry}

\usepackage{amsmath, amsfonts, amssymb}
\usepackage{mathtools}
\usepackage{bm}
\DeclareMathOperator*{\assign}{\doteq}

\DeclareMathOperator*{\argmax}{arg\,max}
\newcommand{\transpose}[1]{{#1}^\intercal}
\newcommand{\Normal}[2]{\mathcal{N}(#1,#2)}
\newcommand{\signoise}{\sigma_{\epsilon}}
\newcommand{\cov}[1]{\mathrm{cov}\left[#1\right]}
\newcommand{\corr}[1]{\mathrm{corr}\left[#1\right]}
\newcommand{\EI}[1]{\mathrm{EI}(#1)}
\newcommand{\MFEI}[1]{\mathrm{MFEI}(#1)}
\newcommand{\GP}[2]{G\!P(#1, #2)}

\newcommand{\R}{\mathbb{R}}

\newcommand{\E}{\mathbb{E}} 

\newcommand{\cA}{\mathcal{A}}
\newcommand{\cD}{\mathcal{D}}
\newcommand{\cM}{\mathcal{M}}
\newcommand{\cP}{\mathcal{P}}
\newcommand{\cS}{\mathcal{S}}

\newcommand{\K}{\bm{K}}
\newcommand{\bx}{\bm{x}}
\newcommand{\bchi}{\bm{\chi}}
\newcommand{\bfopt}{f(\bm{x}^*)}

\newcommand{\bargopt}{\bm{x}^*}
\newcommand{\bargbest}{\bm{x}^+}
\newcommand{\by}{\bm{y}}
\newcommand{\bkappa}{\bm{\kappa}}
\newcommand{\af}{\upsilon}  
\newcommand{\cost}{\lambda}  

\newcommand{\undmax}{\ensuremath{\text{max}}}

\usepackage{booktabs}
\usepackage{multirow}
\usepackage{array}
\usepackage{longtable}

\usepackage{graphicx}
\usepackage{subcaption}
\graphicspath{{images/}}

\usepackage{xcolor}

\usepackage[ruled,vlined]{algorithm2e}

\usepackage[version=4]{mhchem}
\usepackage{siunitx}
\usepackage{longtable,tabularx}
\title{Resource Aware Multifidelity Active Learning\\for Efficient Optimization}
\author[1]{F. Grassi}
\author[1,2]{G. Manganini}
\author[1,3]{M. Garraffa}
\author[1]{L. Mainini}
\affil[1]{United Technologies Research Centre Ireland, Ltd, Cork City, Ireland}
\affil[2]{Gran Sasso Science Institute, L'Aquila, Italy}
\affil[3]{University College Cork, Cork City, Ireland}

\date{}
\begin{document}
\maketitle

\begin{abstract}
Traditional methods for black box optimization require a considerable number of evaluations which can be time consuming, unpractical, and often unfeasible for many engineering applications that rely on accurate representations and expensive models to evaluate. Bayesian Optimization (BO) methods search for the global optimum by progressively (actively) learning a surrogate model of the objective function along the search path. Bayesian optimization can be accelerated through multifidelity approaches which leverage multiple black-box approximations of the objective functions that can be computationally cheaper to evaluate, but still provide relevant information to the search task. Further computational benefits are offered by the availability of parallel and distributed computing architectures whose optimal usage is an open opportunity within the context of active learning. 
This paper introduces the Resource Aware Active Learning (RAAL) strategy, a multifidelity Bayesian scheme to accelerate the optimization of black box functions. At each optimization step, the RAAL procedure computes the set of best sample locations and the associated fidelity sources that maximize the information gain to acquire during the parallel/distributed evaluation of the objective function, while accounting for the limited computational budget. The scheme is demonstrated for a variety of benchmark problems and results are discussed for both single fidelity and multifidelity settings. In particular we observe that the RAAL strategy optimally seeds multiple points at each iteration allowing for a major speed up of the optimization task.
\end{abstract}

\section{Introduction}
\label{sec:introduction} 

\lettrine{O}{ptimization} problems are common in aerospace science and engineering. Practical examples include the design of vehicles, systems and structures, which require the evaluation of disciplinary models and objective functions that are frequently treated as black-box functions.  
Typically, an optimization algorithm operates sequentially by evaluating the objective function at a given point based on its previous evaluations till some stopping criteria is met. When the evaluation of the function is expensive, traditional methods for black-box optimization -- in which a considerable number of evaluations is required -- are poorly suited for such applications. Surrogate Based Optimization (SBO) can significantly improve the efficiency of the optimization procedure: the available information is exhausted and synthetized into a surrogate model to lower the amount of required expensive function evaluations thus saving time, resources and the associated costs~\cite{jones1998efficient,queipo2005surrogate,eldred2006formulations,robinson2008surrogate, forrester2009recent,bhosekar2018advances}.  Efficiency can be further improved in a multifidelity setting, where we have cheaper, but potentially biased approximations to the function that can be used to assist the search of optimal points~\cite{kennedy2000predicting, forrester2007multi,fernandez2016review,park2017remarks, peherstorfer2018survey, beran2020comparison}. 
Within this context, we propose a scheme for resource-aware multifidelity active learning to reduce the computational time and cost associated with the optimization of black-box functions. We aim to achieve this goal through the optimal exploitation of computational budgets (time and computing resources) and of the information contained in the surrogate model (continuously updated while searching for the optimum).

Multifidelity active learning for the optimization of black-box functions has been popularly studied in the Bayesian Optimization (BO) setting~\cite{viana2013efficient,lam2015multifidelity, takeno2019multi}, which consists of two components: (i) a Bayesian statistical model to approximate the objective function, and (ii) an acquisition function to decide where to sample next~\cite{brochu2010tutorial, frazier2018tutorial}. The statistical models are almost invariably Gaussian Processes (GP), for their capability to model arbitrary complex functions, analytical tractability and profitability to estimate uncertainty in a probabilistic framework~\cite{williams2006gaussian, forrester2007multi, kennedy2000predicting, le2014recursive, cutajar2019deep}. The search for the optimum is guided by an acquisition function -- computed on the statistical surrogate model -- which defines a metric for evaluating the next point to sample, balancing the trade-off between a global exploration and a local exploitation of the surrogate. The BO framework for the multifidelity settings combines different information sources (the objective function and its approximations at different levels of fidelity) into a single surrogate model and implements active learning strategies by adaptively sampling from different fidelity levels.

Multifidelity Bayesian Optimization is largely explored in the literature~\cite{viana2014special,guo2018analysis,meliani2019multi, kontogiannis2020comparison}. However, many challenges are still open to the research community.
The optimization of the multifidelity acquisition function is of critical importance for the implementation of an effective active learning strategy, and it may be computationally demanding: in real-world physics-based problems (e.g. the design of aerospace systems and vehicles), the acquisition function is defined over multidimensional domains and subject to non-trivial/non-convex constraints limiting the space of feasible and acceptable solutions~\cite{gardner2014bayesian, hernandez2016general, perrone2019constrained}.
Moreover, the rationale behind the construction and optimization of the acquisition function, at each sampling step, is the balance between exploration and exploitation thrusts: exploitation involves greedily improving over an already good point and exploration is the attempt to gain information about the optimum in under-explored regions. This motivates the interest not only for the point-wise maximization of the acquisition function, but also for its overall form and shape assumed over the entire search domain. This aspect is crucial within an active learning process and contributes to the knowledge acquisition and uncertainty reduction towards the optimization of the black-box function. Finally, BO approaches commonly meet difficulties in optimally exploiting a given computational budget and greedy strategies are usually adopted, which simply maximize the acquisition function point-wise.

Stemming from these open challenges, this paper proposes a scheme for resource-aware multifidelity active learning to assist/inform and accelerate optimization. In particular, we present a computational approach to enable: (i) constraints-aware, space filling sampling; (ii) optimal allocation of available resources at each single step, including leveraging parallel computing architectures at best through the optimal distribution of sample evaluations; (iii) optimally informative multipoint and multifidelity sampling at each step.

To achieve these goals, we formulate the sampling task at each step of the BO as a knapsack problem to select multiple points and allocate resources for their evaluation. Specifically, this means identifying the best candidate locations and the associated fidelity sources in order to maximize the information gain that can be acquired during a parallel evaluation of the objective function, while accounting for the limited computational budget.
Differently from most of the approaches, rather than explicitly optimizing the acquisition function~\cite{wilson2018maximizing}, we evaluate it on a set of feasible points checked beforehand and then consider the problem of selecting an appropriate subset of candidate points with good informative properties, coherently with a knapsack problem approach. By splitting the feasibility check and the points selection tasks, it is possible to fast optimize even complex multifidelity acquisition functions constrained over a non-convex domain.
The knapsack problem is implemented as a mixed-integer linear programming (MILP) model over the candidate points within the feasible domain. The domain is partitioned into strata to capture multiple features of the acquisition function by sampling it in wisely distributed locations~\cite{dambrosio2017milp}. During the (active) learning process the choice of the sampling locations is driven and refined at each step through adaptive discretization techniques. 
The optimized sampling procedure is aware of the computational time budget and of the parallel computing resources available, which are therefore leveraged to balance the trade-off between exploration and exploitation in a principled way. In addition, the optimal use of the available resources for the learning process permits a major contraction of the time required to approach and eventually achieve (or closely approximate) the optimum.

The paper is organized as follows. 
Section~\ref{sec:BO} discusses the setup of the Bayesian Optimization problem and its extension to multifidelity formulations. Section~\ref{sec:raal} introduces the Resource Aware Active Learning scheme (RAAL for short) with formulations for multipoint and multifidelity adaptive sampling. Section~\ref{sec:results} demonstrates the RAAL scheme for the multifidelity optimization of a variety of standard analytical test functions and for classical benchmark problems. Finally, Section~\ref{sec:conclusions} summarizes the concluding remarks.

\section{Optimization framework}
\label{sec:BO}

\subsection{Bayesian Optimization}
Bayesian Optimization (BO) is a class of machine learning techniques for the efficient optimization of expensive black-box functions~\cite{brochu2010tutorial,frazier2018tutorial}. Let consider the constrained optimization problem in the form:
\begin{equation}\label{eq:optimization}
    \min_{\bx \in \cA} f(\bx).
\end{equation}
where $\bx \in \R^d$ is the input, $\cA$ is a feasible set in which it is easy to assess membership, and $f(\bx)\in\R$ is the continuous objective function.
In this context, the term black-box denotes functions that lack of any special structure, as concavity or linearity, or for which derivatives are not known. This is the case in a wide range of applications, such as design of engineering and control systems~\cite{mockus2012bayesian, forrester2008engineering, holicki2020controller}, design of laboratory experiments~\cite{negoescu2011knowledge, packwood2017bayesian}, model calibration~\cite{majda2010quantifying}, reinforcement learning~\cite{lizotte2007automatic, brochu2010tutorial, cutler2014reinforcement}, and hyperparameter tuning of machine learning algorithms~\cite{snoek2012practical, chakraborty2020transfer}. In the following, we will denote $\bfopt$ the solution to problem~\eqref{eq:optimization}, and $\bargopt$ its location. The BO framework consists of two components: a Bayesian surrogate model for modelling the objective function, and an Acquisition Function (AF) for deciding where to sample next. The surrogate models are frequently in the form of Gaussian Processes (GP) that can provide efficient representations of complex functions and characterize model uncertainty in probabilistic frameworks (Section~\ref{sec:GP}). The search for the optimum is guided by an acquisition function defined on the statistical surrogate and defines a metric for evaluating the next point to sample through a continuous trade-off between a global exploration and a local exploitation of the surrogate (Section~\ref{sec:AF}).

\subsubsection{Gaussian processes}
\label{sec:GP}
The main building block of our approach is the Gaussian Process regression~\cite{williams2006gaussian}. Let consider a dataset of $n$ paired input/output observations $\cD_n=\{(\bx_i,y(\bx_i))\}_{i=1}^n$, with $\bx_i\in\R^d$ and $y(\bx_i)\in\R$, generated by the unknown mapping function $y(\bx)=f(\bx)+\epsilon$, where $\epsilon\sim\Normal{0}{\signoise}$ is the measurement noise. The GP regression defines a supervised problem in which we associate to the function $f$ a GP prior having mean 0 and covariance function $\kappa \colon \R^d \rightarrow \R$, such that
\begin{equation}
    f\sim \GP{0}{\kappa(\bx,\bx')}.
\end{equation}
Denoting $\K\in\R^{n\times n}$ the kernel matrix, such that $\K(i,j) = \kappa(\bx_i,\bx_j)$, and $\bkappa_n(\bx)\assign(\kappa(\bx,\bx_1),\dots,\kappa(\bx,\bx_n))$, the predictive distribution of the GP is defined by the mean function $\mu(\bx)$ and the variance function $\sigma^2(\bx)$
\begin{subequations}\label{eq:predictive_gp}
\begin{align}
    \mu(\bx) &= \transpose{\bkappa_n(\bx)} (\K + \signoise\bm{I})^{-1}\by\\
    \sigma^2(\bx) &= \kappa(\bx,\bx) - \transpose{\bkappa_n(\bx)} (\K + \signoise\bm{I})^{-1}{\bkappa_n(\bx)},
\end{align}
\end{subequations}
where $\by\assign\transpose{(y(\bx_1),\dots,y(\bx_n))}$ and $\bm{I}$ the $n$-dimensional identity matrix.

\subsubsection{Acquisition function}
\label{sec:AF}
Once we have a statistical model to represent our belief about the unknown function $f$ given $\cD_n$, we need a sampling strategy or policy for selecting the new query point $\bx_{n+1}$. In Bayesian optimization, the selection strategy utilizes the posterior distribution to guide the search and usually consists in the maximization of a quantity that measures how much information this query will provide, i.e. its expected utility. More formally, the unknown objective function $f$ will be evaluated at $\bx_{n+1}=\argmax_{\bx} \af(\bx\mid\cD_n)$ where $\af(\cdot)$ is the Acquisition Function (AF). Common acquisition functions are the Probability of Improvement (PI)~\cite{kushner1964new}, Expected Improvement (EI)~\cite{jones1998efficient}, entropy Search (ES)~\cite{hennig2012entropy} and Predictive Entropy Search (PES)~\cite{hernandez2014predictive}. The results of this work are obtained using the EI, which, given its analytical tractability and good trade-off between computational cost and accuracy, is the most widely used in the literature~\cite{jones1998efficient}. The Expected Improvement is defined as
\begin{equation}\label{eq:EI}
    \EI{\bx} = \E[\max(f(\bx) - f(\bargbest),0)] =  \begin{cases} 
    (\mu(\bx) - f(\bargbest) - \zeta)\Phi(Z) +  \sigma(\bx) \phi(Z) & \mbox{if }\sigma(\bx)>0\\ 
    0, & \mbox{if } \sigma(\bx)=0 \end{cases}
\end{equation}
where $\mu(\bx)$ and $\sigma(\bx)$ are the predictive in equations in~\eqref{eq:predictive_gp}, $f(\bargbest)$ is the value of the best sample so far and $\bargbest$ is the location of that sample. $\Phi$ and $\phi$ are the cumulative distribution function (CDF) and the probability density function (PDF) of the standard normal distribution, respectively, and $Z$ the standardized improvement
\begin{equation}
    Z = \begin{cases} 
    \dfrac{(\mu(\bx) - f(\bargbest) - \zeta)}{\sigma(\bx)} & \mbox{if }\sigma(\bx)>0\\ 
    0, & \mbox{if } \sigma(\bx)=0. \end{cases}
\end{equation}
The parameter $\zeta$ allows to tune the trade-off between exploration and exploitation, determining the relative importance of the posterior mean $\mu(\bx)$ with respect to the potential improvement in region with high uncertainty, i.e. large $\sigma(\bx)$.

\subsection{Multifidelity Bayesian Optimization}
\label{sec:MFBO}
Multifidelity optimization approaches leverage the availability of analysis models characterized by different levels of fidelity. Typically, high fidelity models consist of ground-truth observations, which are costly to obtain, and/or accurate computer representations of the physics which can be expensive to evaluate. Cheap low-fidelity models may come in various forms: coarser discretizations and resolutions of numerical models, simplified representations which neglect physical effects included in the more expensive high-fidelity models, or approximations through surrogate modeling techniques.

\subsubsection{Multifidelity Gaussian processes}
\label{sec:MFGP}
The Gaussian process regression can be extended to combine different sources of information in a single probabilistic model. For this purpose, let assume that $y^{(1)}(\bx),\dots,y^{(M)}(\bx)$ observation values are available at $M$ different fidelity levels, where $y^{(1)}(\bx)$ is the lowest fidelity and $y^{(M)}(\bx)$ the highest. The training dataset $\cD_n=\{(\bx_i,y^{(m_i)}(\bx_i), m_i)\}_{i=1}^n$ is then composed by the paired input/output observation $(\bx_i, y^{(m_i)}(\bx_i))$, generated by the $m_i$ unknown mapping function $y^{(m)}(\bx)=f^{(m)}(\bx)+\epsilon$, where the measurement noise $\epsilon\sim\Normal{0}{\signoise}$ is assumed to have the same distribution over the fidelities. 
In this setting, the multifidelity Gaussian process regression (MF-GP) can be formulated using an autoregressive scheme~\cite{kennedy2000predicting}, where the lowest fidelity function is characterized by a GP prior $f^{(1)}\sim \GP{0}{\kappa_1(\bx,\bx')}$ with kernel function $\kappa_1\colon\R^{d\times d}\rightarrow \R$, and the higher fidelities are defined recursively as
\begin{equation}\label{eq:argp}
    f^{(m)}(\bx) = \rho f^{(m-1)}(\bx) + \delta^{(m)}(\bx) \qquad m=2,\dots,M
\end{equation}
where $\rho$ is a constant factor that scales the contribution of the preceding fidelity to the following one, and ${\delta^{(m)}(\bx) \sim \GP{0}{\kappa_m (\bx,\bx')}}$ models the bias between fidelities.

The autoregressive formulation implies the following
\begin{equation}\label{eq:markov}
    \cov{f^{(m)}(\bx),f^{(m-1)}(\bx') \bigm| f^{(m-1)}(\bx)} = 0 \qquad \forall \bx\neq\bx',
\end{equation}
which can be interpreted as a Markov property: given the point $f^{(m-1)}(\bx)$, we can learn nothing more about $f^{(m)}(\bx)$ from any other model evaluation $f^{(m-1)}(\bx')$, for $\bx\neq\bx'$~\cite{kennedy2000predicting, o1998markov}.
A kernel function between a pair of samples $\{(\bx_i,y^{(m_i)}(\bx_i), m_i),(\bx_j,y^{(m_j)}(\bx_j), m_j) \}$ can be written as
\begin{equation}
     \kappa((\bx_i,m_i),(\bx_j,m_j)) = \cov{f^{(m_i)}(\bx_i), f^{(m_j)}(\bx_j)}.
\end{equation}

Denoting $\K\in\R^{n\times n}$ the kernel matrix, such that $\K(i,j) = \kappa((\bx_i,m_i),(\bx_j,m_j))$, the predictive distribution of the MF-GP is defined by the predictive mean and variance
\begin{subequations}\label{eq:predictive_mfgp}
\begin{align}
    \mu^{(m)}(\bx) &= \transpose{\bkappa^{(m)}_n(\bx)} (\K + \signoise\bm{I})^{-1}\by\\
    \sigma^{2(m)}(\bx) &= \kappa((\bx,m),(\bx,m)) - \transpose{\bkappa^{(m)}_n(\bx)} (\K + \signoise\bm{I})^{-1}{\bkappa^{(m)}_n(\bx)},
\end{align}
\end{subequations}
where $\bkappa_n(\bx)\assign\transpose{(\kappa((\bx,m),(\bx_1,m_1)),\dots,\kappa((\bx,m),(\bx_n,m_n)))}$ and $\by\assign\transpose{(y^{(m_1)}(\bx_1),\dots,y^{(m_n)}(\bx_n))}$.

\subsubsection{Multifidelity acquisition function}
\label{sec:MFAF}
The availability of multiple fidelity levels poses a new challenge for the Bayesian Optimization: not only we have to determine the location of the new sample to evaluate, but also the most convenient fidelity level to query. Different approaches can be found in the literature, as Multifidelity Expected Improvement (MFEI)~\cite{huang2006sequential}, Multifidelity Predictive Entropy Search (MFES)~\cite{zhang2017information} or Multifidelity Max-value Entropy Search~\cite{takeno2019multi}. For consistency, our formulation is based on the MFEI, which preserves the good properties of its single fidelity counterpart. The MFEI (or Augmented EI)~\cite{huang2006sequential} is defined as
\begin{equation}\label{eq:MFEI}
    \MFEI{\bx,m} = \E[\max(f^{(M)}(\bx) - f^{(M)}(\bargbest),0)] \alpha_1 (\bx, m) \alpha_2 (\bx, m),
\end{equation}
where the first term is simply the EI evaluated at the highest fidelity, therefore can be simply derived from~\eqref{eq:EI}. The utility functions $\alpha_1 (\bx, l)$ and $\alpha_2 (\bx, l)$ are defined as
\begin{align}
\alpha_1 (\bx, m) &= \corr{f^{(m)}(\bx), f^{(M)}(\bx)} \\
\alpha_2 (\bx, m) &= 1 - \dfrac{\signoise}{\sqrt{\sigma^{2(m)}(\bx) +  \signoise^2}},
\end{align}
where $\alpha_1(\bx, m)$ is designed to discount the utility when a lower fidelity evaluation is considered, whereas $\alpha_2(\bx, m)$ takes into account the stochastic nature of the unknown function $f$ due to the presence of noise, therefore for deterministic problems is $\alpha_2(\bx, l)=1$.  The function $\alpha_1(\bx, m)$ has been chosen to be straightforward to compute under the assumptions of the GP regression, and it holds $\alpha_1(\bx, M)=1$ and, for deterministic problems, $\alpha_1(\bx, m)=0$ if $(\bx,y^{(m)}(\bx), m)\in\cD_n$. For a more detailed analysis of the MFEI, please refer to~\cite{huang2006sequential}.

\section{Resource-Aware Active Learning}
\label{sec:raal}

The computational improvements introduced by the adoption of multifidelity surrogates can be further enhanced by the optimal use of parallel or distributed computing architectures. This section introduces our Resource-Aware Active Learning algorithm (RAAL, for short) that leverages the availability of multiple sources of information at different levels of fidelity in conjunction with the possibility to distribute the evaluations across multiple computational resources (CPUs) in parallel.
Section~\ref{sec:RAALOverview} outlines the main steps of the RAAL algorithm. We then discuss the two elements representing the core of the approach: the multipoint exploration/exploitation of the AF, in Section~\ref{sec:af_discretization}, and the optimization procedure for the multipoint multifidelity seeding in Section~\ref{sec:optimization_routine}.
In the following, for the sake of compactness, we write $[n]$ for $\{1,\dots,n\}$.

\subsection{Overview}
\label{sec:RAALOverview}

\begin{algorithm}[t]
\SetAlgoLined
\DontPrintSemicolon
\KwIn{Feasible set $\cP \subset \R^d$, multifidelity objective function $f()$, priors $GP(0,\kappa_m(\bx,\bx'))$, parallel CPUs budget $\{\beta_g\}_{g=1}^G$}
\KwOut{(Approximate) Solution $\bargopt$ to~\eqref{eq:optimization}}

\BlankLine

Select initial points $\cS_0 \subset \cP$ with associated fidelities $\cM_0$\;
$t \leftarrow 0$, $B \leftarrow 0$\;
\Repeat{Stop criteria is met on iterations $t<t_\undmax$ or used budget $B<B_\undmax$}{
    \tcp{Parallel function evaluation}
    $\cD_{n_t}  \leftarrow$ Evaluate $f()$ at points $\bx \in \cS_t$ at fidelities $m \in \cM_t$ on $G$ parallel CPUs\;
    $\cP \leftarrow \cP \setminus \cS_t$\;
    
    \tcp{Learn/Update surrogate model}
    Update posteriors $\mu^{(m)}, \sigma^{2(m)}$ in~\eqref{eq:predictive_mfgp} based on $\cD_{n_0:n_t} \assign \{\cD_{n_0},\dots,\cD_{n_t}\}$\;
    
    \tcp{Multipoint multifidelity seeding}
    \Begin($\forall \bx_i \in \cP$){
        Discretize points: $\bchi_i \leftarrow \Xi(\bx_i,\xi_t)$ \tcc*[r]{Section~\ref{sec:af_discretization}}
        Evaluate the AF: $\af_{i}^{(m)} \leftarrow \af(\bx_i,m), \, \forall m \in [M]$ 
    }
    $\cS_{t+1}, \cM_{t+1} \leftarrow$ Solve~\eqref{eq:milp} based on
    $\{\bchi_i\}_{i=1}^N$, $\{\af_{i}^{(m)}\}_{i=1,m=1}^{N,M}$, $\{\cost^{(m)}\}_{m=1}^M$, $\{\beta_g\}_{g=1}^G$
    \tcc*[r]{Section~\ref{sec:optimization_routine}}
    
    \tcp{Next iteration}
    $B \leftarrow B + \sum_{m \in \cM_t} \cost^{(m)}$\;
    Update $\xi_t$ by \eqref{eq:xi_update}\;
    $t \leftarrow t + 1$\;
}
\Return{Point $\bx$ with minimum $y^{(M)}(\bx)$ over the whole dataset $\cD_{n_0:n_t}$}
\caption{Resource-Aware Active Learning (RAAL) Algorithm}
\label{alg:raal}
\end{algorithm}

In the conventional BO, one point is selected at each iteration and evaluated at a prescribed fidelity level. This information is then used to learn/update the surrogate model and the associated AF, before the next point selection can be made. In the remaining of this paper, this conventional BO is referred to as {\em sequential} BO. Differently, the RAAL scheme samples multiple points across different fidelities at each iteration. In addition, the RAAL scheme optimally allocates the computational resources available to take most advantage of parallel computing and/or distributed computing architectures.

We describe here the main steps of the RAAL strategy (Algorithm~\ref{alg:raal}).
We start from an initial set of feasible points $\cP = \{\bx_1, \dots, \bx_N\} \subset \R^d$ that can be assembled through any Design of Experiment (DOE) procedure, for instance a Latin Hypercube Design (LHD), and then remove the unfeasible points $\bx \notin \cA$. At the first iteration $t=0$, a subset $\cS_0$ of $n_0$ points is selected together with a set of fidelity levels $\cM_0$ and used to run the first evaluations of the function $f$, and hence obtain the dataset $\cD_{n_0}$ employed to learn an initial surrogate model $f^{(M)}$ and the associated AF, as described in Section~\ref{sec:BO}. At this point, similarly to the sequential BO, the RAAL algorithm optimizes the AF, but with the notable difference of selecting a subset of points $\cS_1 \subset \cP \setminus \cS_0$ together with the associated fidelity levels $\cM_1$. At the next iteration $t=1$, parallel computational resources are used to built simultaneously the dataset $\cD_{n_1}$, where $n_1 > 1$, and, similarly to the sequential BO, the optimization loop is executed based on the augmented dataset $\cD_{n_0:n_1} \assign \{\cD_{n_0}, \cD_{n_1}\}$. As it will be showed in the numerical results of Section~\ref{sec:results}, the seeding of more than one point at each BO iteration significantly speeds up the overall computational time, mitigating the impact of the main bottleneck in the BO process, i.e. the evaluation of the black-box function $f$.
The process is then iterated till a maximum number of iterations $t_\undmax$ or a maximum computational budget $B_\undmax$ is reached.

The RAAL multipoint selection comes with a number of favourable properties.
As explained in more details in Section~\ref{sec:af_discretization}, the optimization of the AF -- which may be complex, high-dimensional and multifidelity -- is accomplished by evaluating it point-wise at points in $\cP$ and picking those points that cumulatively maximize the function over the search domain. By doing so it is possible to handle even complex constraints on the design space, since their feasibility is checked beforehand and not during the AF optimization. Besides, the selected points can be chosen to achieve a tunable exploration/exploitation trade-off of the AF, combining space-filling characteristics with selective exploitation of the AF shape.
Another important aspect of the multipoint seeding is that the sampled points maximize the usage of the computational resources available in terms of the computational burden required to evaluate the points at the selected fidelity levels. The aim is to make the most out of the parallel resources in order to reduce the impact of the function evaluations on the BO iterations, also by properly allocating the different evaluation tasks to the parallel CPUs.

Summarizing, the $n_{t+1}$ points in the set $\cS_{t+1}$ and their associated fidelities $\cM_{t+1}$, selected at each iteration $t$ of the RAAL algorithm for the maximization of the AF, (i) are feasible with respect to design constraints, (ii) are well-distributed and have tunable space-filling properties, and (iii) maximize the usage of available computational resources.
Section~\ref{sec:af_discretization} will describe how the points should be processed in order to take into consideration a trade-off between AF exploration and exploitation; Section~\ref{sec:optimization_routine} will describe the multipoint multifidelity optimization routine.

\subsection{Optimal exploration/exploitation of the Acquisition Function in multipoint scenario}
\label{sec:af_discretization}

The optimization of the AF in the RAAL multipoint scenario relies on a tunable exploitation and exploration of the AF, which will be actively employed in the optimization procedure of Section~\ref{sec:optimization_routine} for the selection of multiple points at each BO iteration.

When a feasible set of points $\cP$ in the AF domain is given, the maximization (i.e. exploitation) of the AF itself can be done by simply evaluating the function in these points for each required level of fidelity, obtaining a set of values $\af(\bx_i,m)$, with $\bx_i \in \cP$ and $m \in [M]$, and then picking the highest value.
In a standard BO scheme, this is an optimization-by-evaluation procedure that can be especially suitable when the numerical optimization of the AF is particularly challenging, mainly due to the presence of complex and non-convex feasibility constraints. While still convenient in a parallel-BO scheme, some attention must be paid: indeed, a greedy selection of only the best points would lead to oversampling the AF in a close neighborhood of the optimum, without consequently providing much additional information instead of picking the single optimum and, in fact, wasting computational resources.

A better strategy can be devised if we include aspects not only related to the exploitation, but also to the exploration of the AF.
One way to explore the AF (and hence the domain of the original objective function) is to use experimental design techniques. A very common approach in this field is the LHD, which divides each dimension $j$ of a $d$-dimensional space into $E_j$ equispaced levels (also known as strata or bins), where each level contains exactly one point in the design.
In the literature (see~\cite{dambrosio2017milp} for a brief survey) there are different criteria to evaluate the goodness of an LHD configuration.
A popular criterion is the $L_p$-discrepancy with respect to the uniform distribution, which measures the difference between the empirical distribution of a set of points and the multivariate uniform distribution over the same domain. Such uniform discrepancy is a multidimensional property that is difficult to evaluate, however computing the discrepancy along each one-dimensional projection is a much simpler task. Notice that, for example, the defining property of a LHD is that each one-dimensional projection has low uniform discrepancy along that dimension.

\paragraph{Uniform gridding}
If we take a purely exploration point of view, our measure of the goodness of a set of points is related to the difference with the \textit{ideal} one-dimensional distributions of the points along the $d$ axes. First we divide each continuous domain $j \in [d]$ into a pre-specified number of bins $E_j$ with the mapping $\Xi:\R \to \{0,1\}^{E_j}$, and then project each point onto the $d$ axes:

\begin{equation}\label{eq:uniform_discretization}
(\Xi_j(x))_e \assign
    \begin{cases}
        1 & \text{if } x \in \left[ L_j + (e-1)\frac{U_j-L_j}{E_j}, e\frac{U_j-L_j}{E_j} \right] \\
        0 & \text{otherwise}
    \end{cases},
\end{equation}
where $e \in [E_j]$ and $[L_j,U_j]$ represent the interval of the $j$-th domain.
Then, each vector $\bx_i \in \mathcal{P}$ is mapped onto an extended vector $\Xi(\bx_i) \assign (\Xi_1(x_{i,1}),\dots,\Xi_d(x_{i,d})))$. If the points in a set $\cS \subset \cP$ were uniformly distributed, the projection along each axis would look like a univariate uniform distribution. Given equispaced strata of each variable $j$, each stratum should contain the same number of points, i.e $|\cS|/E_j$.
In Section~\ref{sec:optimization_routine} we will see how to implement this in a proper optimization problem.

\begin{figure}[h!t]
	\centering
	\includegraphics[width=0.8\columnwidth]{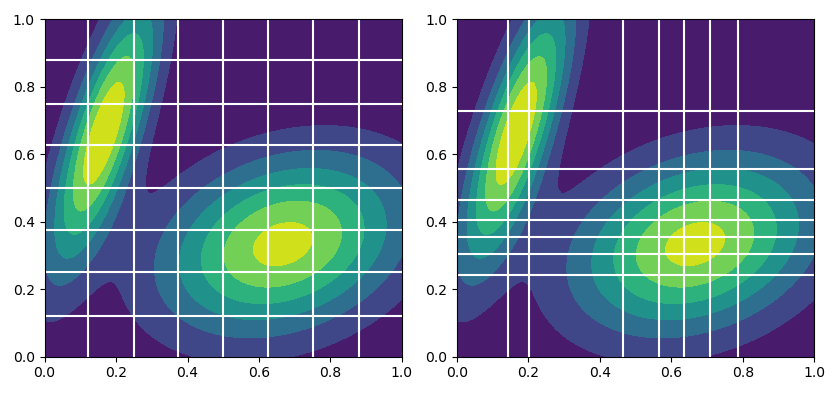}
	\caption{An example of adaptive gridding techniques on a synthetic 2D AF profile with with $E_j=8$, $j=[2]$. Left figure represent a uniform grid obtain when $\xi=0$, while the right figure shows the impact of $\xi=0.8$ in the generation of a finer grid around AF modes.}
	\label{fig:adaptive_grid}
\end{figure}

\paragraph{Adaptive gridding}
The discretization technique of Equation \eqref{eq:uniform_discretization} produces a uniform grid without taking into consideration the shape of the AF, since the points in $\cP$ are discretized into partitions of equal length/width and recall as close as possible the measure of discrepancy from the uniform distribution. 
In order to balance the pure exploration given by the uniform gridding with the function exploitation of the AF, we propose a AF-weighted discretization technique.
In place of transforming the points in $\cP$ on a purely geometric basis, the AF values $\af(\bx_i,m)$ are used to guide their discretization. First, for each dimension $j \in [d]$, the points with an AF value smaller than a predefined quantile level $\xi \in (0,1)$ are removed. Subsequently, quantiles $Q_j^e$ are computed on the AF values of the remaining points, where $Q_j^e$ is the $e$-th $E_j$-quantile of the points along the dimension $j$, with $0<e<E_j$.
Figure~\ref{fig:adaptive_grid} offers an illustrative example: on the left we see the case with $\xi=0$, resulting in a uniform grid, while on the right we see the impact of selecting a $\xi>0$, which results in a more refined grid around the modes of the AF.

The parameter $\xi$ can also be automatically updated during the iterations of the RAAL algorithm, so as to favour exploration at the beginning of the algorithm, and foster exploitation as we get close to the depletion of the available computational budget. Formally, said $\xi_\undmax$ the maximum allowed value for the parameter $\xi$, the update rule can be specified as
\begin{equation} \label{eq:xi_update}
    \xi_t = \xi_\undmax \left(1- \exp{\left(- \eta_\xi \dfrac{B}{B_\undmax - B}\right)}\right)
\end{equation}
where $\xi_t$ is the parameter value at iteration $t$, $B$ is the current budget over the total $B_\undmax$ and $\eta_\xi$ is the learning rate.

Once defined the quantiles for each dimension, each point in $\cP$ is projected onto the $j$ axis by the mapping
\begin{equation}\label{eq:quantile_discretization}
(\Xi_j(x,\xi))_e \assign
    \begin{cases}
        1 & \text{if } x \in \left[ Q_j^e, Q_j^{e+1} \right] \\
        0 & \text{otherwise}
    \end{cases}
\end{equation}
where $e \in [E_j]$, and $Q_j^0 \assign L_j$, $Q_j^{E_j} \assign U_j$.
In this case we highlighted the dependence on the parameter $\xi$ because, contrary to~\eqref{eq:uniform_discretization}, the width of the bins is adjusted depending on the AF values distribution using the quantile-based method. This leads to a finer grid resolution in those areas where the AF has higher values, and therefore where it is more likely to sample more informative points.
Similarly to the previous case, each vector $\bx_i \in \cP$ is finally mapped onto an extended vector $\Xi(\bx_i, \xi) \assign (\Xi_1(x_{i,1}, \xi),\dots,\Xi_d(x_{i,d}, \xi))$. Section~\ref{sec:optimization_routine} will show hot to implement proper constraints such that the projection along each axis looks like a univariate uniform distribution to obtain well distributed points over the defined grid.

\subsection{Multipoint multifidelity seeding}
\label{sec:optimization_routine}

When multiple computational resources are available at each iteration of the BO loop, we face the problem of how to best utilize these resources to gain as much information as possible from the current surrogate model and its related AF.
Section~\ref{sec:af_discretization} already discussed the proposed strategies for the multipoint maximization of the AF, in the direction of balancing exploitation and exploration. This section describes how these strategies can be embedded in an optimization program that takes into consideration three main aspects characterizing our multipoint multifidelity seeding, namely:
\begin{enumerate}
\item the maximization of the usage of the computational resources available and their optimal allocation for the evaluation of the objective function $f$ at different fidelity levels;
\item the optimal exploitation and exploration of the AF;
\item a sampling strategy compatible with the recursive GP model used to build the surrogate (Section~\ref{sec:MFGP}).
\end{enumerate}

From an high level perspective, the seeding routine is implemented as a knapsack problem, where the candidate points and the relative fidelity sources are selected so that the information acquired during a parallel evaluation of the objective function is maximized, and the computational load is less than or equal to the available parallel resources.
Consequently, the optimization problem takes four groups of input parameters: the points to be selected $\bx_i$, transformed and `decoded' into their categorical version $\bchi_i \assign \Xi(\bx_i)$, thanks to the discretization procedures~\eqref{eq:uniform_discretization} or~\eqref{eq:quantile_discretization}; the $\af_{i}^{(m)}$ values of the AF evaluated at points $\bx_i \in \mathcal{P}$ and fidelities $m \in [M]$; the computational cost $\cost^{(m)}$ of evaluating the objective function $f$ at fidelity $m \in [M]$; the computational resources $\beta_g$ of each single computational unit $g \in [G]$, where $G$ is the number of available CPUs, to be allocated for the evaluation of $f$ at the fidelity levels that will be selected.

The decision variables that allow for the selection of the next points and their fidelities to be evaluated (i.e. the sets $\cS_{t+1}$ and $\cM_{t+1}$ in Algorithm~\ref{alg:raal}) are arranged into two groups: variables $p_{i,g}^{(m)}$ equal $1$ iff the point $\bx_i$ at fidelity $m$ is chosen and assigned to the computational unit $g \in [G]$; variables $q_i$ equal $1$ iff the point $\bx_i$ is chosen from $\cP$, independently from the fidelity level. The variables $q_i$ are used in combination with the discretized data $\bchi_i$ to select a set of points such that each stratum of the discretized grid is represented, in order to measure a (scaled) discrepancy with respect to the uniform distribution. It is worth reminding that, according to the discretization procedure chosen from Section~\ref{sec:af_discretization}, such measure enforces the property of weighted well-distributed points, hence balancing the exploration/exploitation of the AF itself over the selected points.
On the other hand, variables $p_{i,g}^{(m)}$ determine the fidelity level $m$ and the computational unit $g$ assigned to point $\bx_i$ to minimize the waste of resources, accounting for the specific computational cost $\cost^{(m)}$ associated to the fidelity level $m \in [M]$.

The optimization routine is finally formulated as the following Mixed Integer Linear Programming (MILP) problem:
\begin{subequations}
\label{eq:milp}
\begin{align}
    \min_{p_{i,g}^{(m)}, q_i} \quad
    & \Upsilon\left( \sum_g^G \beta_g - \sum_{i,m,g}^{N,M,G} \cost^{(m)} p_{i,g}^{(m)}\right) - \sum_{i,m,g}^{N,M,G} \af_{i}^{(m)} p_{i,g}^{(m)} \nonumber \\
    s.t \quad
    \label{eq:milp:b}
    & \sum_i^N \chi_i^e q_i \leq 1 \quad \forall e \in [E] \\
    \label{eq:capacity}
    & \sum_{i,m}^{N,M} \cost^{(m)} p_{i,g}^{(m)} \leq \beta_g \quad \forall g \in [G] \\
    \label{eq:milp:var_coherence1}
    & \sum_{m,g}^{M,G} p_{i,g}^{(m)} \leq (M*G)q_i\quad \forall i \in [N] \\
    \label{eq:milp:var_coherence2}
    & q_i \leq \sum_{m,p}^{M,G} p_{i,g}^{(m)}  \quad \forall i \in [N] \\
    \label{eq:milp:cpu}
    & \sum_g^G p_{i,g}^{(m)} \leq 1 \quad \forall i \in [N], m \in [M]\\
    \label{eq:milp:gp_cnst}
    & \sum_g^G p_{i,g}^{(m)} \leq \sum_g^G p_{i,l}^p\quad \forall m \geq l, \,l,m \in [M]\\
    & p_{i,g}^{(m)},q_i \in \{0,1\} \quad i \in [N], m \in [M], g \in [G] \nonumber 
\end{align}
\end{subequations}
where $E=\sum_{j=1}^d E_j$ is the total number of bins used in the processing of data $\bchi_i$.

The MILP objective induces a hierarchical order in the optimization of the two goals of maximizing the usage of resources and maximizing the AF: the former is prioritized over the latter through the weighting factor $\Upsilon$ set such that $\Upsilon>\!>\sum_{i,m}^{N,M} \af_{i}^{(m)}$. With constraints~\eqref{eq:milp:b} we impose that at most one point can be chosen that belongs to each bin $e \in [E]$, hence enforcing the well-distributed property described in Section~\ref{sec:af_discretization}. Constraints~\eqref{eq:capacity} guarantee that the capacity of each computational unit is not violated when the evaluations of the objective function are assigned. Logical interdependence between the groups of variables $q_i$ and $p_{i,g}^{(m)}$ are is imposed by constraints~\eqref{eq:milp:var_coherence1} and~\eqref{eq:milp:var_coherence2}, while~\eqref{eq:milp:cpu} assure that a single point cannot be evaluated on more than one CPU. Finally, the fidelity interdependence for the construction of the coherent auto-regressive GP model (described in Section~\ref{sec:BO}) is implemented by~\eqref{eq:milp:gp_cnst}.

\section{Experiments}
\label{sec:results}

In this section we present numerical experiments demonstrating the performances of the RAAL algorithm compared to a standard sequential BO scheme (i.e. using a single CPU), for which we use a set of benchmark problems. In the following, we first describe the experimental setup followed in all the experiments, including the benchmarks description and the tests configurations; then we move to discuss the results obtained in both single and multifidelity versions of the benchmark problems. 

We implemented the RAAL algorithm, its statistical models and acquisition functions in Python 3.7.3, leveraging functionality from the Emukit toolkit~\cite{emukit2019}, while the MILP Optimization Routine of Section~\ref{sec:optimization_routine} was implemented with PuLP~\cite{mitchell2011pulp}, a linear programming modeler written in Python, and solved by means of COIN CLP/CBC~\cite{lougee2003common}.

\subsection{Experimental setup}
\label{sec:exsetup}

\begin{figure}[t]
	\centering
	\includegraphics[width=1\columnwidth]{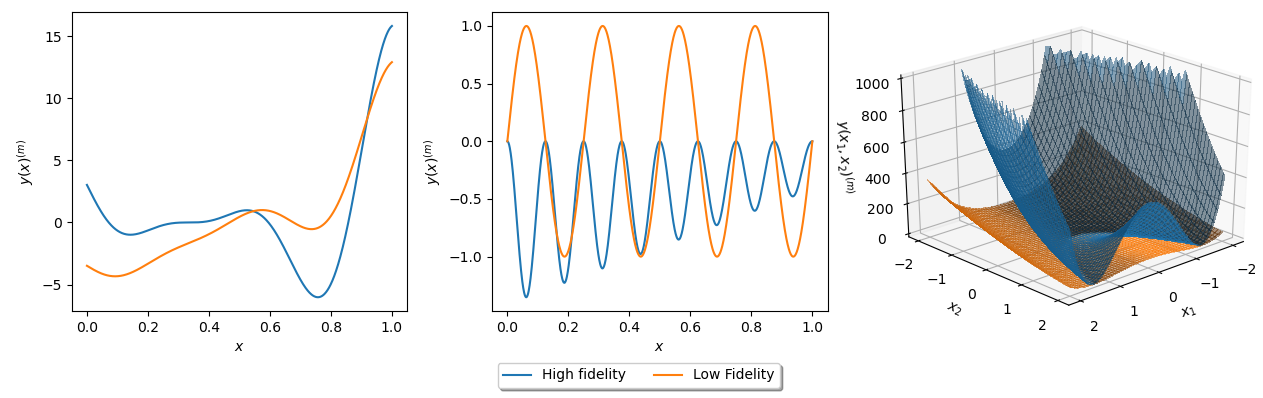}
	\caption{Analytical benchmark objective functions and their low fidelity alternatives: Analytical Test 1 (left),  Analytical Test 2 (center), and Analytical Test 3 with $d=2$ (right).}
	\label{fig:benchmarks}
\end{figure}

We conducted experiments on a variety of popular benchmark problems to test the efficiency and robustness of the proposed approach against the standard sequential BO, either in single (SF) and multifidelity (MF) settings. The benchmark functions were selected to exemplify different types of correlations among the fidelity levels, described in the following. Consistently with the already used notation, we denote $y^{(m)}(\bx)$ the objective functions, and sort the fidelities in an increasing order $m=1,\dots,M$. Accordingly, $M$ is the representation at the highest-fidelity and is considered the reference ground-truth.
\paragraph{Analytical Test 1}
The first benchmark is the popular Forrester function~\cite{forrester2008engineering}, one of the most common analytical benchmark in the literature. It is a 1-dimensional nonlinear function over the domain $[0,1]$, defined as   
\begin{equation}
y^{(2)}(\bx) = (6\bx - 2) ^ 2  \sin(12  \bx - 4),
\end{equation}
with $\bargopt\simeq0.727549$ and $\bfopt\simeq-6.02074$.
Its low fidelity level is given by the linear mapping
\begin{equation}
y^{(1)}(\bx) = 0.5  y^{(2)}(\bx) + 10 (\bx - 0.5).
\end{equation}

\paragraph{Analytical Test 2}
The second benchmark is a sinusoidal squared 1-dimensional function~\cite{cutajar2019deep}, with domain in the interval $[0,1]$. The high fidelity function is defined as
\begin{equation}
y^{(2)}(\bx) = (\bx - \sqrt{2}) (y^{(1)}(\bx)) ^ 2,
\end{equation}
which is a non linear function of the low fidelity variant, given by
\begin{equation}
y^{(1)}(\bx) = \sin(8 \pi \bx).
\end{equation}
Its ground truth solution to problem~\eqref{eq:optimization} is $\bfopt\simeq-1.35201$ at $\bargopt \simeq 0.0619147$.

\paragraph{Analytical Test 3}
The third benchmark problem is the $d$-dimensional Rosenbrock function~\cite{rosenbrock1960automatic}, a non-convex function with domain in the interval $[-2,2]^d$ and defined as
\begin{equation}
    y^{(2)}(\bx) = \sum_{i=1}^{d-1} (1-x_i)^2 + 100(x_{i+1} - x_i)^2 \qquad \mathrm{where} \; \bx=[x_1,\dots,x_d]\in\R^d.
\end{equation}
The global minimum $\bfopt=0$ lies in a narrow, parabolic valley and is located at $\bargopt = [1,\dots,1]^d$.
The low fidelity observations are given by a linear mapping defined as~\cite{bryson2017unified}:
\begin{equation}
    y^{(1)}(\bx) = \dfrac{  y^{(2)}(\bx) - 4.0 - \sum_{i=1}^{d} 0.5 x_i } { 3.0 + \sum_{i=1}^{d} 0.25 x_i}.
\end{equation}

Figure~\ref{fig:benchmarks} illustrates the three analytical objective functions, together with their low fidelity alternatives. In the remaining of the paper ``SF Test $\ell$'' denotes the optimization of the $\ell$-th Analytical Test problem in a single fidelity setting, where just the highest fidelity level is considered. Conversely, ``MF Test $\ell$'' indicates the optimization of the $\ell$-th Analytical Test problem in a multifidelity setting, considering all the available fidelity levels as available sources of information.

For all the numerical results, same initial conditions were imposed to each algorithm configuration: an identical initial set $\cS_0$, with cardinality dictated by the specific benchmark application, was selected randomly from the feasible set $\cP$, drawn quasi-randomly via LHD over the feasible domain of each benchmark.
We also allocated, for each experiment, the same maximum total computational budget $B_\undmax$ to both the sequential BO and the RAAL algorithm, i.e. the highest level of fidelity can be evaluated the same number of times
Finally, for all the analytical benchmarks, we set a unitary cost to the maximum fidelity level and a fractional cost to all the lower fidelity level, according to the following rule
\begin{equation}
\lambda^{(m)} = \begin{cases}
1 & \text{if } m = M, \\
\dfrac{1}{5(M-m)} & \text{if } m < M.
\end{cases}
\end{equation}
In the RAAL algorithm, each available CPU was assigned with a computational budget capable of running a single evaluation of the objective function at the maximum fidelity level.

In the following results we report the Root Squared Error between the optimal solution computed at each step and the known global optimum (minimum) of the high fidelity function of each benchmark. 
The error is plotted as a function of the iterations used by each algorithm, for which we allocate the same total computational budget to fairly compare the results. This metric is directly related to the execution time taken by the sequential and parallel BO, given that the computational overhead of choosing the next information source and sample is omitted, as it is negligible compared to invoking an information source in real-world applications.
All the numerical experiments were randomized over 20 runs, from different initial sets $\cS_0$: all diagrams reports the median values (solid lines) together with all the other observations falling in the interval between the 25-th and 75-th percentiles (shaded areas).
The hyperparameters of the kernel and mean functions of the GP surrogate models were optimized via Maximum Likelihood Estimation~\cite{forrester2007multi,forrester2009recent}.

\subsection{Single fidelity results}
\label{sec:SFresults}

This section discusses the results observed for the single fidelity version (SF) of the artificial benchmarks; this set of experiments permits to investigate the impact of different parameter values of the RAAL algorithm on its performances, namely the accuracy and the speed of convergence to the known optimum.
In particular, we focus our attention on different grid resolutions $E$ and different values of the learning rate parameter $\eta_\xi$ while varying the number of available CPUs (that is the number of points that can be evaluated simultaneously).

\begin{figure}[t]
	\centering
	\includegraphics[width=0.99\columnwidth]{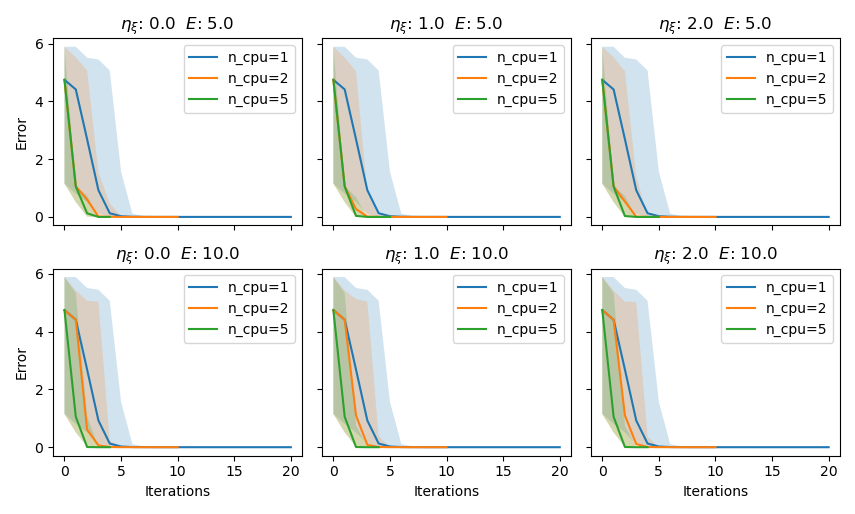}
	\caption{SF Test 1. Comparison between sequential BO and RAAL BO with 2 CPUs and 5 CPUs: impact of the number of bins $E$ in the search point discretization, and of the learning rate $\eta_\xi$ in the adaptive grid. Tests were run with a maximum budget of $B_\undmax=20$.}
	\label{fig:sf_forrester}
\end{figure}

Figure~\ref{fig:sf_forrester} shows the results on the SF Test 1 benchmark, for which we chose an initial DOE of $n_0=2$ points. We run the tests with $B_\undmax=30$, and all the different parameters combinations resulting from two discretization levels $E=5$ and $E=10$ and three learning rates $\eta_\xi=0,1,2$.
First of all we can see how the RAAL algorithm achieves better results than the sequential BO in terms of convergence speed: the RAAL algorithm takes 2.5 iterations on average to reach the optimum in case 5 CPUs are employed, whereas the sequential BO takes on average 5 iterations. Similar results are obtained for all the parameter settings of the grid discretization, that is the number of bins $E$ and the learning rate $\eta_\xi$.
From these experiment, the learning rate seems not to have significant impact onto the convergence speed. This holds for the simple case of SF Test 1, whose AF shape may be fairly simple to be captured and exploited.

\begin{figure}[t]
	\centering
	\includegraphics[width=0.99\columnwidth]{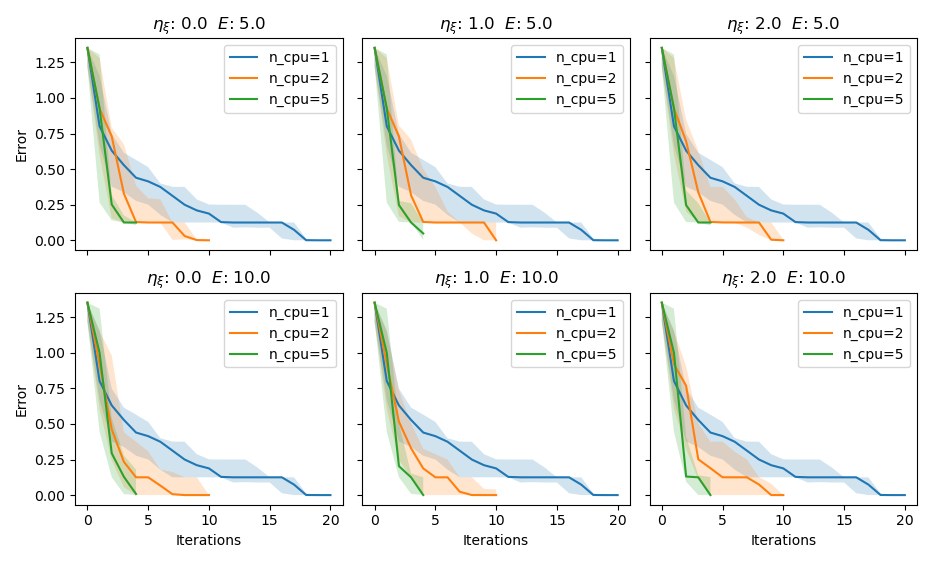}
	\caption{SF Test 2. Comparison between sequential BO and RAAL BO with 2 CPUs and 5 CPUs: impact of the number of bins $E$ in the search point discretization, and of the learning rate $\eta_\xi$ in the adaptive grid. Tests were run with a maximum budget of $B_\undmax=30$.}
	\label{fig:sf_nonlinearA}
\end{figure}

We move now to investigate the impact of different numbers of bins $E$ in the discretization grid and of the learning rate $\eta_\xi$ on the SF Test 2. For this test we used $n_0=2$ points as initial DOE and a maximum computational budget of $B_\undmax=30$. Interestingly, Figure~\ref{fig:sf_nonlinearA} shows similar results to Figure~\ref{fig:sf_forrester}: the parallel selection of multiple points leads to a significantly improved convergence speed, without compromising the performances in terms of accuracy. While the sequential BO takes 20 iterations to reach the optimum, the RAAL algorithm takes 10 iterations with 2 CPUs and less than 5 iterations with 5 CPUs. Another important aspect regards the resolution of the grid: increasing the resolution of the grid by doubling the number of bins from $E=5$ (top row) to $E=10$ (bottom row) helped the RAAL achieve a faster convergence and avoid local optima, represented by those plateau in the algorithm iterations. It may be deduced that, given the high multimodality of SF Test 2, the RAAL algorithm can benefit from a higher number of bins in the search grid, which allows a finer search and the movement from one local optima to another, till reaching the true objective function optimum.
Lastly, higher learning rates degrade the performance in presence of local optima, both with a coarse and a finer grid. The main reason is that the adaptation yields denser sampling in the proximity of the peaks of the acquisition function, therefore mitigating the exploration thrusts of the uniform gridding which would be beneficial to skip out of local minima. This is confirmed by the behaviour observed for the uniform gridding, for which we record shorter stagnation at the local minimum.

\begin{figure}[h!t]
	\centering
	\includegraphics[width=0.8\columnwidth]{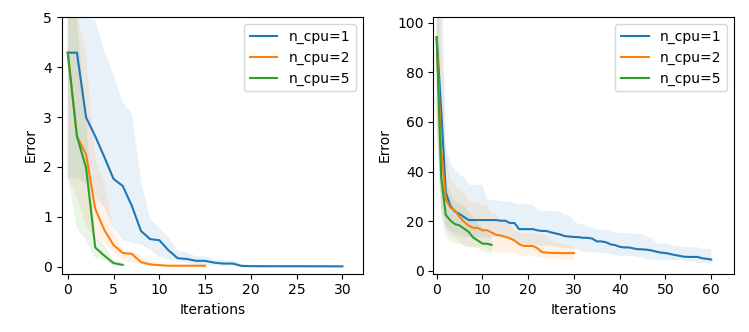}
	\caption{SF Test 3, with $d=2$ (left) and $d=4$ (right). Comparison between sequential BO and RAAL BO with 2 CPUs and 5 CPUs. RAAL parameters are $E_j = 5$, $j \in [d]$ and $\eta_\xi=0$.}
	\label{fig:sf_rosenbrock24}
\end{figure}

An additional analysis was carried out on the multidimensional domain of the SF Test 3 (Rosenbrok), where we set uniform gridding for the entire optimization procedure ($\eta_\xi=0$) and $E_j = 5$ bins for each dimension. Investigations were conducted for different dimensionality of the problem, namely for $d=2$ and $d=4$.
Also in this scenario, the RAAL BO outperforms the sequential BO in terms of convergence speed, even if with $d=4$ the sequential BO achieves, on average, a slightly better accuracy.  A possible reason for this is that, ideally, the cardinality of the set of points to evaluate at each iteration increases with the dimensionality of the domain to sample.

\subsection{Multifidelity results}
In this section we describe the results for the multifidelity (MF) version of the analytical benchmarks and discuss the impact of the RAAL parameters. In addition, the outcomes are compared to the single fidelity experiments, in order to verify whether similar considerations can be drawn. Similarly to the single fidelity case, we investigate the impact of different parameters of the RAAL algorithm, that is the numbers of bins $E$ in the discretization grid and the learning rate $\eta_\xi$. In particular we investigate their role for the 2 CPUs and the 5 CPUs architectures.

\begin{figure}[h!t]
	\centering
	\includegraphics[width=0.55\columnwidth]{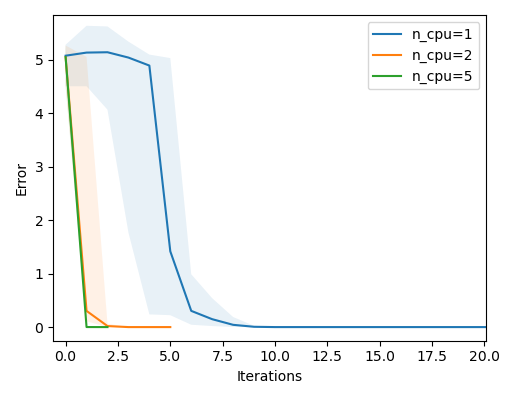}
	\caption{MF Test 1. Comparison between sequential BO and RAAL BO with 2 CPUs and 5 CPUs with $B_\undmax=10$. RAAL parameters are $E=5$ and $\eta_\xi=0$.}
	\label{fig:mf_forrester}
\end{figure}

Figure~\ref{fig:mf_forrester} reports the results obtained for the benchmark MF Test 1, with a maximum computational budget $B_\undmax=10$. Here we use a uniform grid of $E=5$ bins and establish an initial set of 5 and 2 points for the low and high fidelity levels, respectively. 
The multipoint selection of the RAAL BO permits to sensitively accelerate the convergence to the optimum; this already emerges when 2 CPUs only are available. Moreover, the parallel selection of different points reduces the variability of the results across the experiments, that is, the proposed multipoint and multifidelity seeding enhances the robustness of the BO scheme with respect to a sequential approach.

\begin{figure}[t]
	\centering
	\includegraphics[width=0.99\columnwidth]{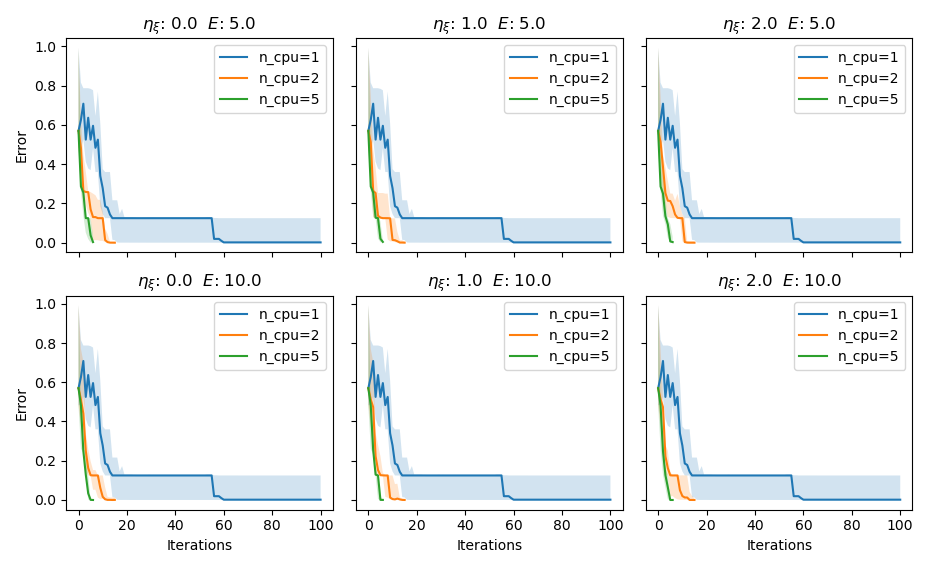}
	\caption{MF Test 2. Comparison between sequential BO and RAAL BO with 2 CPUs and 5 CPUs: impact of the number of bins $E$ in the search point discretization, and of the learning rate $\eta_\xi$ in the adaptive grid. Tests were run with a maximum budget of $B_\undmax=20$.}
	\label{fig:mf_nonlinearA}
\end{figure}

We run the MF Test 2 with $B_\undmax=20$ for all the settings of the algorithmic parameters resulting from two discretization levels $E=5$ and $E=10$, and three learning rates $\eta_\xi=0$, $\eta_\xi=1$, and $\eta_\xi=2$. Similarly to the MF Test 1, the initial DOE consists of 5 and 2 points for the low and high fidelity, respectively.  Figure~\ref{fig:mf_nonlinearA} shows that, also in this second multifidelity benchmark, the multipoint selection dramatically accelerates the search of the optimum which in many cases can be found in only 5 iterations when exploiting 5 CPUs.
Furthermore, it is worth noticing that the RAAL algorithm performs better in this MF Test 2 rather than in its single fidelity version SF Test 2.
The comparison of Figures~\ref{fig:sf_nonlinearA} and~\ref{fig:mf_nonlinearA} reveals that the RAAL algorithm always achieves the global optimum of the benchmark function in the MF setting, whereas it does not manage the same in the SF case, when 5 CPUs are used. In fact, the access to a lower fidelity and less costly representation of the objective function allows the RAAL algorithm to sample more points and to better explore the search space, which turns out to be very useful for highly multimodal problems of this kind.
For what concerns the effect of different discretization levels, we can observe that a coarse $E$ leads to better optimization performance when we have a smaller number of CPUs, whereas a finer grid allows to achieve faster convergence when the number of CPUs is higher. This is related to trade-off between the exploration and the exploitation thrusts: the combination of a lower number of CPUs with a finer grid is too unbalanced towards the exploitation, whereas a coarse grid paired with a high number of CPUs (and therefore samples per iteration), biases the optimization in favour of the exploration. Lastly, differently from what observed for the single fidelity settings, increments of the learning rate $\eta_\xi$ do not have any significant impact on the convergence history of the multifidelity implementation of benchmark Test 2.

\begin{figure}[h!t]
	\centering
	\includegraphics[width=0.99\columnwidth]{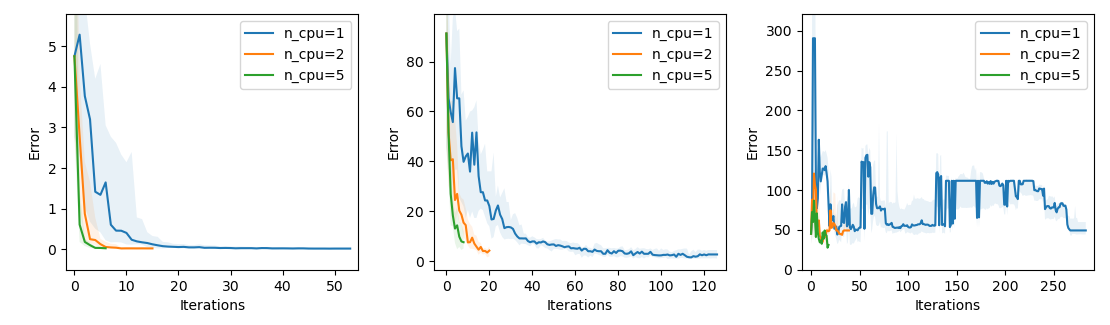}
	\caption{MF Test 3, with $d=2$ (left), $d=4$ (center), and $d=8$ (right), with $B_\undmax=30$, $B_\undmax=40$, and $B_\undmax=80$, respectively. Comparison between sequential BO and RAAL BO with 2 CPUs and 5 CPUs. RAAL parameters are $E_j=5$, $j \in [d]$, and $\eta_\xi=0$.}
	\label{fig:mf_rosenbrock}
\end{figure}

Eventually, experiments are reported for the mutidimensional benchmark Test 3 (Rosenbrock function) in the multifidelity scenario.
Investigations have been conducted for $d=2,4$ and $8$ dimensional domains with different maximum budget $B_\undmax=30$, $40$, and $80$, respectively allocated; a uniform gridding is adopted to discretize each dimension with $E_j = 5$ bins.

Similarly to what observed for the single fiedelity experiment, the results recorded for the multifidelity settings (Figure~\ref{fig:mf_rosenbrock}) demonstrate the faster convergence speed of the RAAL BO, which is particularly impressive in the highest dimensional domain of $d=8$: the parallel multipoint selection of the RAAL algorithm leads to a smaller final error with respect to the true optimum, which was achieved in a little fraction of the iterations taken by the sequential BO.

\section{Concluding Remarks}
\label{sec:conclusions}
In this work we proposed a novel multipoint and multifidelity Bayesian Optimization (BO) scheme, with the objective of accelerating the optimization of expensive-to-evaluate black box functions.
Our Resource Aware Active Learning (RAAL) algorithm is able to maximize the information gain to acquire at each step of the underlying BO methodology by seeding multiple points and the associated fidelities while optimally allocate parallel/distributed computational resources available for their evaluation.
The core of the algorithm is the seeding procedure, implemented as a mathematical programming problem, which leverages in a principled way the computational time budget and parallel resources available to balance the trade-off between exploration and exploitation of the Acquisition Function (AF), leading a major speed up in the iterative optimization task.
Another main characteristic of the RAAL algorithm is its general formulation, which can scale to any finite number of fidelities, handle any statistical model and deal with any AF. This should guarantee a wide applicability of the approach, without limiting its validity to any specific BO-related setting.

The performances of the approach were empirically evaluated on a number of well-known analytical benchmarks available in the literature, with non-linear and multimodal characteristics, tested with two fidelity levels for demonstration purposes. The results obtained for all the numerical experiments reveal a significant speed up of the RAAL algorithm in solving the optimization problem with respect to a standard BO scheme, where the AF is optimized and sampled in only one point.
Interestingly enough, the RAAL achieves even better performances in  multifidelity scenarios, demonstrating the ability to take full advantage of the lower fidelity and cheaper-to-evaluate approximation of the objective function in seeding more points and hence better explore the search domain at each algorithm iteration.

As potential extension of this work, we are currently investigating different opportunities.
First, numerical results should be extended to physics-based applications and problems, for an additional validation of the approach for physics-based multidomain use cases.
Another worthwhile investigation may regard the use of opportunely extended Multipoint Acquisition Functions, explicitly formulated so as to maximize the information gain either at the same BO iteration or over a look-ahead on future iterations, recalling a Dynamic Programming approach. Some attempts are already available in the literature, but they only focus on the single fidelity scenario.
Lastly, a potential advancement of the algorithm can be its adaptation to the so-called Constrained Bayesian Optimization, where the objective function has to be optimized in presence of expensive-to-evaluate feasibility constraint, which usually involve the formulation of modified Acquisition Functions.

\section*{Acknowledgments}
This work was supported by the IDA Center of Excellence in Cyber Physical Systems Grant No. 176474 under the Industrial Development Agency (Ireland) program.

\bibliographystyle{IEEEtran}
\bibliography{biblio.bib}

\end{document}